\documentclass[final]{cvpr}

\usepackage{times}
\usepackage{epsfig}
\usepackage{graphicx}
\usepackage{epstopdf}
\usepackage{amsmath}
\usepackage{amssymb}
\usepackage{booktabs}
\usepackage{multirow}
\usepackage[ruled]{algorithm2e}

\usepackage[pagebackref=true,breaklinks=true,colorlinks,bookmarks=false]{hyperref}

\def\cvprPaperID{8559} % *** Enter the CVPR Paper ID here
\def\confYear{CVPR 2021}

\begin{document}

%%%%%%%%% TITLE
\title{Box Re-Ranking: Unsupervised False Positive Suppression for \\
Domain Adaptive Pedestrian Detection}

\author{Weijie Chen$^{1,2}$, Yilu Guo$^2$, Shicai Yang$^2$, Zhaoyang Li$^2$, Zhenxin Ma$^2$, \\
Binbin Chen$^2$, Long Zhao$^2$, Di Xie$^2$, Shiliang Pu$^2$, Yueting Zhuang$^1$\\
$^1$Zhejiang University, Hangzhou China\\
$^2$Hikvision Research Institute, Hangzhou, China\\
{\tt\small \{chenweijie5,guoyilu5,yangshicai,lizhaoyang5,mazhenxin\}@hikvision.com}\\
{\tt\small {\{chenbinbin8,zhaolong11,xiedi,pushiliang.hri\}@hikvision.com},yzhuang@zju.edu.cn}
}
\maketitle

%%%%%%%%% ABSTRACT
\begin{abstract}
False positive is one of the most serious problems brought by agnostic domain shift in domain adaptive pedestrian detection. However, it is impossible to label each box in countless target domains. Therefore, it yields our attention to suppress false positive in each target domain in an unsupervised way. In this paper, we model an object detection task into a ranking task among positive and negative boxes innovatively, and thus transform a false positive suppression problem into a box re-ranking problem elegantly, which makes it feasible to solve without manual annotation. An attached problem during box re-ranking appears that no labeled validation data is available for cherry-picking. Considering we aim to keep the detection of true positive unchanged, we propose box number alignment, a self-supervised evaluation metric, to prevent the optimized model from capacity degeneration. Extensive experiments conducted on cross-domain pedestrian detection datasets have demonstrated the effectiveness of our proposed framework. Furthermore, the extension to two general unsupervised domain adaptive object detection benchmarks also supports our superiority to other state-of-the-arts.
\end{abstract}

%%%%%%%%% BODY TEXT
\section{Introduction}

\begin{figure}[t]
    \centering
    \includegraphics[width=1.0\columnwidth]{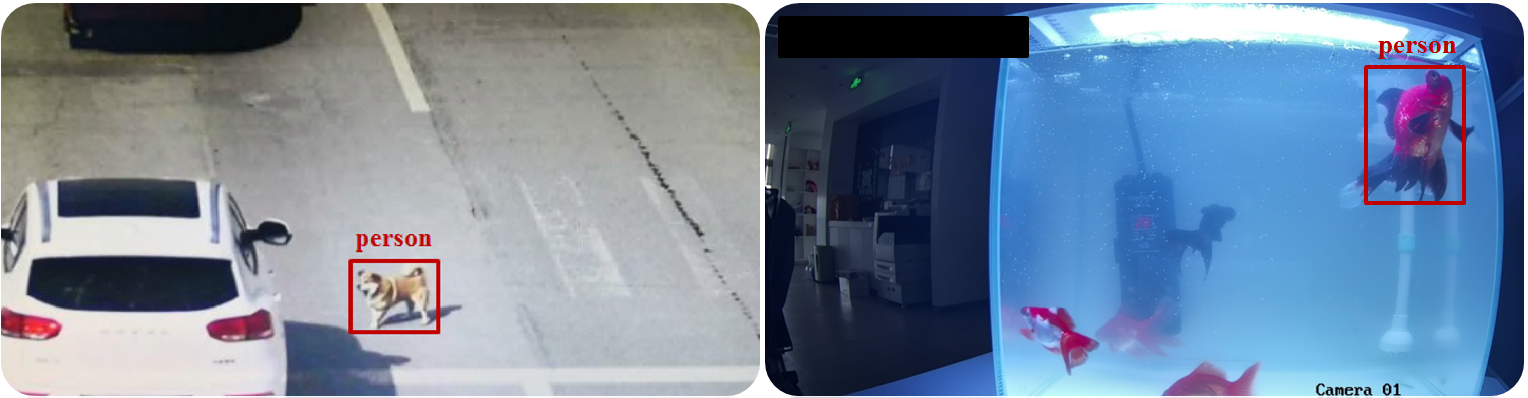} 
    \caption{Two cases of false positive in pedestrian detection. Left (from social news): a dog is false-detected as a pedestrian. Right (self-collected data): Indoor video surveillance for strangers' break-in, but what alerting is always the goldfishes.}
    \label{case}
\end{figure}

Pedestrian detection \cite{song,graininess,bibox,xin2020,shanshan2020,Wei2020} is a very important problem in the field of computer vision and sees many practical applications, such as video surveillance, autonomous driving and so on. Generally, a well-performed pedestrian detection model is trained on a very large-scale annotated images (source domain data), and then embedded in edge devices to process the images sampled from practical scenarios (target domain data). This paradigm makes it a very challenging problem due to agnostic domain shift, namely different context between source domain and target domain. False positive is one of the most serious problems during cross-domain pedestrian detection. Two examples are shown in Fig.\ref{case}. The left one is from social news, where a dog is false-detected as a pedestrian. The right one is an indoor video surveillance scenario for monitoring the break-in of strangers. However, what alerting is always a crowd of goldfishes raised in the room, making the users very annoying. Actually, false positive samples are totally diverse in different target domains. However, it is impossible to label each target domain data manually, which motivates us to study unsupervised false positive suppression for cross-domain pedestrian detection.

\begin{figure*}[t]
    \centering
    \includegraphics[width=1.8\columnwidth]{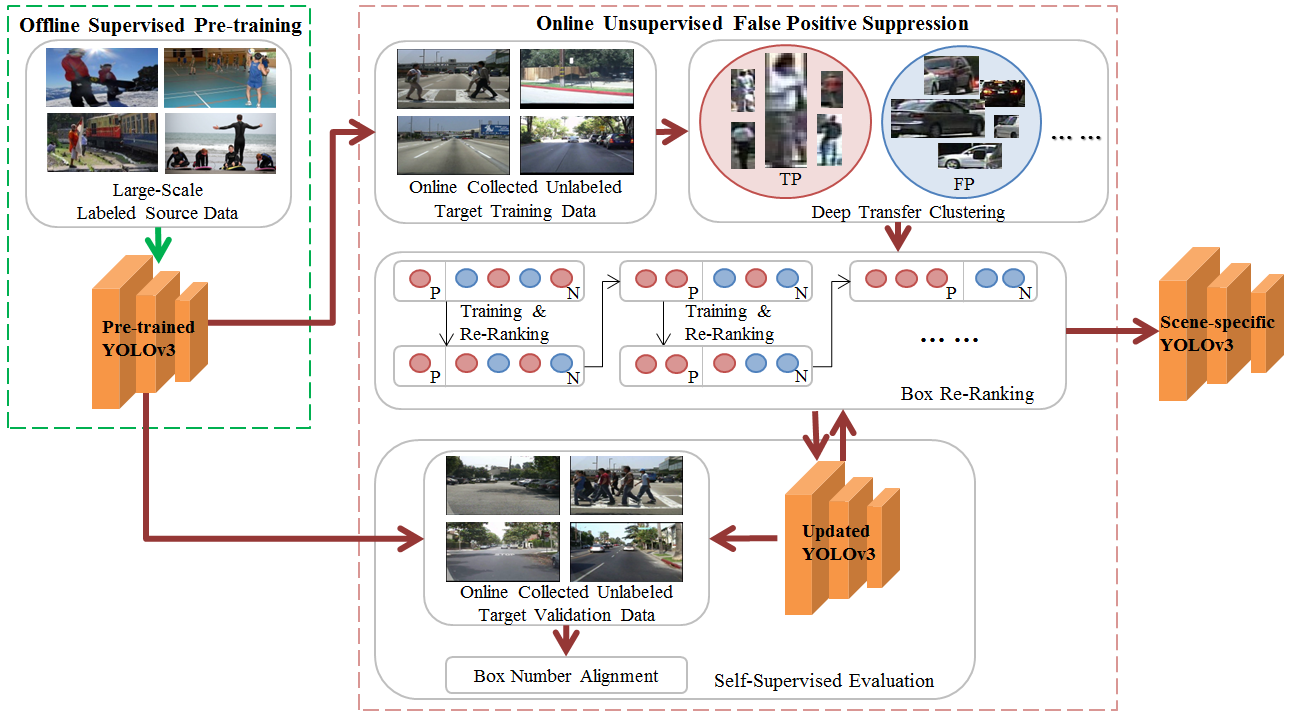} 
    \caption{Unsupervised false positive suppression for domain adaptive pedestrian detection via \emph{box re-ranking}. Only source domain pre-trained model instead of source data is provided for optimization. The target data is collected online, and splited into training and validation parts for \emph{box re-ranking} and self-supervised evaluation, respectively.}
    \label{pipeline}
\end{figure*}

As mentioned above, a very large-scale source annotated data is a prerequisite to pre-train a well-performed pedestrian detection model. However, during domain adaptive optimization, such large-scale source data is hard to store in edge devices due to data transmission as well as data privacy-protection. Especially, source data is unable to suppress the unseen false positive, which is an open-set problem and can only find the answer from target domain. In the right image of Fig.\ref{case} detected by YOLOv3 \cite{yolov3} trained on MS COCO dataset \cite{2014Microsoft}, goldfish is never seen in MS COCO dataset, which is the main reason leading to false positive. Therefore, source domain data is inaccessible in this paper. Only a pre-trained pedestrian detection model and unlabeled target data collected online are provided for unsupervised false positive suppression.

The most challenging problem lies in no labels are available in target domain. The only supervision cue is the detected bounding boxes attached with their classification confidence from a source domain pre-trained model. Sorting the confidence from higher to lower scores, an object detection task can be viewed as a ranking task, which aims to optimize the positive and negative boxes in a good order. In other words, the occurrence of false positive actually behaves as a wrong order with true positive (or false negative). In this way, a false positive suppression problem can be modeled into a \emph{box re-ranking} problem elegantly.

Empirically, the boxes with extremely high positive confidence are likely to be real positive boxes, especially when pre-trained model is trained on a very large-scale dataset collected from numerous source domains. These boxes can provide prior knowledge about true positive and act as seed positive boxes to rectify the order of the remaining ones. Initially, the seed positive boxes and the remaining ones are directly labeled as positive and negative for fine-tuning, and then the confidence of the remaining ones is re-predicted for re-ranking. Note that treating the remaining boxes as negative is a very critical step in our framework to suppress false positive. The positive candidates in the remaining boxes tend to achieve higher confidence than others after training. Recursively, the top-K positive candidates are removed from the remaining boxes and join into the positive counterpart to repeat the training process. We name the entire repetition as \emph{box re-ranking}. To reduce the complexity, deep transfer clustering \cite{DTC} is the first time to use in an object detection task and pack the boxes of the entire dataset with similar semantic information into a cluster. The boxes in a cluster are re-ranked as a whole, simplifying the granularity of \emph{box re-ranking} from an instance level to a cluster level. Also, since pure seed positive boxes are the preliminaries to rectify the order of the remaining ones, label denoising for seed positive boxes is also an extra benefit from deep transfer clustering. 

Another question followed by \emph{box re-ranking} is when to terminate the repetition without using labeled validation data. Since we hope to keep the detection capacity of true positive unchanged after false positive suppression, we propose \emph{box number alignment}, a self-supervised evaluation metric, for cherry-picking during \emph{box re-ranking}. Specifically, the detected box number should keep unchanged before and after \emph{box re-ranking}. Actually, in many practical scenarios, the false positive are usually occurred along with false negative, which means the more false positive are suppressed, the more false negative will be mined back. Overall, the entire pipeline of our proposed framework is shown as Fig.\ref{pipeline}. 

Extensive experiments are conducted on four pedestrian detection adaptation tasks built with five public pedestrian detection datasets. To further verify our proposed method, we collect a more challenging pedestrian detection dataset for false positive suppression. Also, to compare with other related work, we extend our method to two general domain adaptive object detection tasks. Comprehensive experimental results and the corresponding analysis demonstrate the effectiveness of our proposed framework. It can suppress false positive by a large margin in an unsupervised way without declining the detection capacity of true positive. In summary, our main contributions are listed as follows:

\begin{itemize}
\item To the best of our knowledge, this is the first work aiming to suppress false positive for domain adaptive pedestrian detection in an unsupervised way.
\item An object detection task is modeled into a ranking task innovatively, and a false positive suppression problem is turned into a \emph{box re-ranking} problem elegantly, making it feasible to solve in an unsupervised way.  Box number alignment, a self-supervised evaluation metric, is proposed for cherry-picking during \emph{box re-ranking}.
\item Five datasets are constructed to validate the effectiveness of our proposed unsupervised false positive suppression framework, one of which is our self-collected dataset. The extension to two general object detection adaptation tasks also support our superiority to other state-of-the-arts.
\end{itemize}

%------------------------------------------------------------------------
\section{Unsupervised False Positive Suppression}

It is extremely difficult to assign an exactly right pseudo label to each box in an unsupervised settings. To bypass this challenge, we model a false positive suppression problem into a \emph{box re-ranking} problem. Also, to prevent the degeneration of detection capacity, \emph{box number alignment} is proposed as a self-supervised evaluation metric to cooperate with \emph{box re-ranking}. 

\subsection{Prerequisites}
\label{prerequisites}
Before introducing our proposed unsupervised false positive suppression framework, three important prerequisites have to emphasize:

\begin{itemize}
\item The bounding boxes generated by the pre-trained model are directly applied as ground-truth for box regression training, no matter positive or negative boxes, since the location error is much weaker than classification error in object detection tasks as proved by \cite{borji2019empirical}.

\item Reliable positive boxes are the seed prerequisite for our proposed framework to drive box re-ranking. Usually, those with extreme high positive confidence predicted from source domain pre-trained model are very likely to be real positive boxes. This condition is usually satisfied especially when the pre-trained model is trained on a very large-scale dataset collected from numerous source domains. Beyond this, deep transfer clustering is adopted for further label denoising and keeping the purity of seed positive boxes.

\item In most practical scenarios, false positive are occurred along with false negative. After our proposed \emph{box re-ranking}, the confidence position among false positive and false negative are swapped, which indirectly suppresses false positive while mining back a corresponding part of false negative.
\end{itemize}

\subsection{Box Re-Ranking}
As mentioned above, we select seed positive boxes with very high confidence (greater than a given threshold $h$, 0.95 by default in this paper), while the remaining ones are directly treated as negative, which can be formulated as:
\begin{equation}
\{(b, c)\}=M_s(\mathcal{X})
\label{e1}
\end{equation}
\begin{equation}
\mathcal{Y}=\left\{
\begin{aligned}
\{b\}_P & \,\,\, {\rm if} \,\,\, c \textgreater h \\
\{b\}_N & \,\,\, {\rm if}\,\,\, c \leq h
\end{aligned}
\right.
\label{e2}
\end{equation}
where $M_s$ is a given pre-trained model and $\mathcal{X}$ denotes target training dataset. Exploiting $M_s$ on $\mathcal{X}$, we can initially achieve bounding box set $\{(b, c)\}$ where $b$ denotes a detected box attached with corresponding confidence $c$. $\mathcal{Y}$ is the initialized label set including positive part $P$ and negative part $N$ divided from $\{(b, c)\}$ by a very high confidence threshold $h$.
\begin{equation}
M_t \leftarrow \mathcal{L}_{conf}(M_s | \mathcal{X} ; \mathcal{Y}) + \lambda \mathcal{L}_{loc}(M_s | \mathcal{X} ; \mathcal{Y})
\label{e3}
\end{equation}
Using the initial pseudo labels $(\mathcal{X}, \mathcal{Y})$, the source domain pre-trained model $M_s$ is fine-tuned into $M_t$ as shown in Eqn.\ref{e3}. The objective loss function is a weighted sum of the confidence loss (conf) and the localization loss (loc). As metioned in \ref{prerequisites}, we directly use initially-generated boxes as ground-truth for box regression training. Usually, if the positive part is pure enough, the false positive will be well-suppressed. But it is quite easy to encounter the degeneration of true positive. To avoid this situation, more positive boxes should be mined back from the negative part elaborately.
\begin{equation}
\{(b, c)\}_t=M_t(\mathcal{X})
\label{e4}
\end{equation}
Adopting the optimized model $M_t$, the pre-generated bounding boxes $\{b,c\}$ are re-predicted to update their confidence. To prevent the catastrophic forgetting of bounding box localization, the ground-truth of localization always keeps unchanged. In this way, we use the initial bounding box $\{(b, c)\}$ to match the re-predicted ones $\{(b, c)\}_t$. Let $d_{ij}=\{1,0\}$ be an indicator for matching the $i$-th initial bounding box to the $j$-th re-predicted box. $d_{ij}=1$ only if the IoU of this pair is the greatest one and the IoU should be greater than a given threshold (0.3 by default in this paper). So the confidence update process is:
\begin{equation}
c_i=\sum\limits_{j}d_{ij} \cdot c_{t\_j} {\rm where}\,\,\, c_i \in \{(b,c)\}, c_{t\_j} \in \{(b, c)\}_t
\label{e5}
\end{equation}
After confidence update, the bounding box $\{b\}$ are re-ranked from higher to lower scores. The boxes in negative part with similar appearance to positive part will tend to achieve higher positive confidence $N_{top}$. This process is actually similar to PU-Learning \cite{PUL}. We update the label $\mathcal{Y}$ by removing $N_{top}$ from $N$ and adding $N_{top}$ to $P$:
\begin{equation}
\begin{aligned}
P = &P \cup N_{top} \\
N = &N \setminus N_{top} \\
\mathcal{Y} = &\{P, N\}
\end{aligned}
\label{e6}
\end{equation}
After updating $\mathcal{Y}$, Eqn.\ref{e3}-\ref{e5} are repeated recursively until achieving the best performance. 

\subsection{Box Packing via Deep Transfer Clustering}
\begin{figure}[t]
    \centering
    \includegraphics[width=1.0\columnwidth]{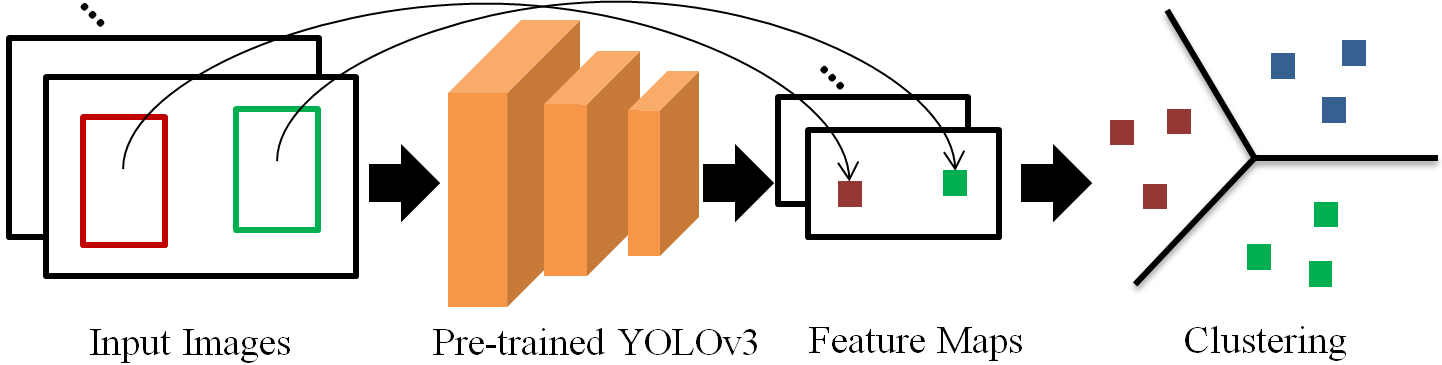} 
    \caption{Deep transfer clustering. We map the center of each box to the feature map of the last second layer of three FPN heads in YOLOv3, and extract the corresponding feature of all boxes in the entire dataset for clustering. }
    \label{clustering}
\end{figure}

It is too redundant to re-rank the boxes in an instance level, since no matter true positive and false positive usually appear frequently in a single target scene. To further discretize the granularity of \emph{box re-ranking}, the boxes with similar semantic feature are packed into a cluster via deep transfer clustering:
\begin{equation}
\min_{C\in \mathbb{R}^{d\times k}} \frac{1}{N} \sum\limits_{n=1}^{N}\min_{y_n\in\{0,1\}^k}||f_s(b_n)-\mathcal{C}y_n||_{2}^{2} \,\, {\rm s.t.}\,\, y_n^\intercal\textbf{1}_k=1
\label{e7}
\end{equation}
where $f_s(b_n)$ means we map the center of the $n$-th box to the feature map $f_s$ generated by $M_s(\cdot)$ and extract the corresponding feature of the box for clustering. $f_s(b_n)\in \mathbb{R}^{d}$ and $d$ is the channel size of $f_s$. $C\in \mathbb{R}^{d\times k}$ is the cluster centroid matrix, $k$ is the cluster number and $y_n$ is the cluster assignment for box $b_n$. $N$ is the box number of the entire dataset. 

As shown in Fig.\ref{clustering}, we use YOLOv3 \cite{yolov3} as an example since we mainly use YOLOv3 in this paper. We concatenate the last second layer of three FPN \cite{Lin2017Feature} heads for feature extraction, and directly map the center of the boxes to the corresponding positions in the feature map. By default, we use $k$-means for boxes clustering. Without specific statement, we over-cluster the boxes into $100$ clusters in this paper. Note that the extracted features include semantic as well as scale and aspect ratio information, since they are used for classification and box regression. Scale and aspect ratio information is beneficial for clustering, since the boxes with similar semantic information usually share similar scale and aspect ratio.

After clustering, the boxes in the same cluster are re-ranked together as a whole via averaging the confidence in a cluster.
\begin{equation}
c_i=\frac{1}{|\Omega|} \sum\limits_{n\in \Omega}c_n \,\,\, {\rm where} \,\,\, \Omega=\{j\}, y_j=y_i
\label{e8}
\end{equation}

Concretely speaking, the boxes whose averaging confidence in the corresponding cluster is greater than $h$ are selected as seed positive boxes. During box re-ranking, the candidate positive boxes are added from the negative part to the positive part cluster by cluster.

An extra benefit of deep transfer clustering is label denoising. A reliable assumption is that the boxes with similar feature should share a similar confidence score consistently. Clustering makes the confidence of each box more robust via voting, avoiding accidently high confidence brought by domain noise and achieving more reliable seed positive boxes. The outlier boxes of each cluster are annotated as ignored labels during optimization.

\subsection{Self-Supervised Evaluation}

In an unsupervised setting, no ground-truth labels are available in target data, which means no annotated validation data is available for cherry-picking during \emph{box re-ranking}. Actually, this is also the main challenge of all unsupervised learning tasks. In this paper, we will introduce a self-supervised evaluation metric \emph{box number alignment} to be an alternative of annotated validation data for false positive suppression tasks.

A main objective during false positive suppression is that the detection capacity of true positive should not be declined. Hence, we aims at keeping the number of detected boxes over a confidence threshold unchanged before and after \emph{box re-ranking}. Actually, in practical application, a confidence threshold $o$ is usually pre-set to output the detected boxes (Note that $o$ is usually smaller than $h$. In practical, $o$ is usually set according to the False Positive Per Image on source domain validation set.). So the box number in the unlabelled validation set are countable and can be used as an evaluation metric for unsupervised false positive suppression. It can be formulated as:
\begin{equation}
\min i \,\,\, {\rm s.t.} \,\,\, C(M_{t\_i}(\mathcal{X}^{`}), o)\ge C(M_s(\mathcal{X}^{`}), o)
\label{e9}
\end{equation}
where $M_s$ and $M_{t\_i}$ denote the pre-rained model in source domain and the $i_{th}$ optimized model in target domain during \emph{box re-ranking}. $\mathcal{X}^{`}$ is an unlabelled validation set in target domain, and $o$ is a confidence threshold to count the output box number after NMS \cite{Neubeck2006EfficientNS}. $C(\cdot)$ separately represent the function of box counting. In general, with our proposed \emph{box number alignment}, the more false positive are suppressed, the more false negative are mined back.

\subsection{Pipeline Overview}
\begin{algorithm}[t]
\label{alg}
\caption{Unsupervised False Positive Suppression}
\LinesNumbered
\KwIn{Pre-trained model $M_s$, unlabeled target data $\mathcal{X}$ and $\mathcal{X}^{`}$ for training and validation, a very high confidence threshold $h$ to select seed positive boxes,  and an output confidence threshold $o$ for box number alignment.}
\KwOut{Optimized model $M_t$}
Bounding boxes initialization $\leftarrow$ Eqn.\ref{e1}\;
Box packing via deep transfer clustering $\leftarrow$ Eqn.\ref{e7}\;
Update $c$ via averaging score in each cluster $\leftarrow$ Eqn.\ref{e8}\;
Generate pseudo label $\mathcal{Y}$ by $h$ $\leftarrow$ Eqn.\ref{e2}\;
Train and get $M_{t\_0}$ $\leftarrow$ Eqn.\ref{e3}\;
Box number counting for evaluation $N_s=C(M_s(\mathcal{X}^{`}), o)$\;
\If{$C(M_{t\_0}(\mathcal{X}^{`}), o)\geq N_s$}{
    return $M_{t\_0}$ \,\,\,\,\,\,$\leftarrow$ Eqn.\ref{e9}\;
}
\For{$i=1$ \textbf{to} $n$}{
	Update $c$ via re-prediction $\leftarrow$ Eqn.\ref{e4},\ref{e5}\;
	Update $c$ via clustering result $\leftarrow$ Eqn.\ref{e8}\;
	Box re-ranking by sorting $c$\;
	Update $\mathcal{Y}$ by removing the boxes in a cluster with highest average score from negative to positive part $\leftarrow$ Eqn.\ref{e6}\;
	Train and get $M_{t\_i}$ $\leftarrow$ Eqn.\ref{e3}\;
    \If{$C(M_{t\_i}(\mathcal{X}^{`}), o)\geq N_s$}{
		return $M_{t\_i}$ \,\,\,\,\,\,$\leftarrow$ Eqn.\ref{e9}\;
		%break\;
    }
}
return $M_{t\_n}$
\end{algorithm}

The entire pipeline of our proposed unsupervised false positive suppression method is shown in Fig.\ref{pipeline} and Algorithm.\ref{alg}. In one sentence, we carry out \emph{box re-ranking} in the cluster level until meeting the condition of \emph{box number alignment} to avoid detection degeneration.

%------------------------------------------------------------------------
\section{Deeper Understanding with Related Work}

\paragraph{Unsupervised Domain Adaptive Pedestrian Detection}  The most related work to our paper is from Liu \emph{et al.} \cite{liu2016unsupervised}, which learns scene-specific pedestrian detectors for target domains in crowded scenes without manual annotation (Actually, they had used labeled validation set in target domain.). They optimized the model domain-adaptively via randomly sampling negative instances from source domain and meanwhile annotating positive instances with high confidence from target domain. From this perspective, their method cannot solve the problem of false positive suppression, since their method cannot fully exploit the negative samples in target domain. Our motivations are totally different which exactly raises great differences between the methods. Actually, domain adaptive object detection can be viewed as an open-set problem. The negative instances from different domains are usually different. And some negative instances in target domains may not be ever seen by source domain. Therefore, the negative instances should be sampled from target domain for training. However, to correctly annotate positive or negative label to each boxes in target domain is a very crucial challenge. So we model the unsupervised false positive suppression problem into a \emph{box re-ranking} problem to bypass this challenge without accessing any source domain data and labeled target domain validation data. Other related work like \cite{guan2019unsupervised,my2018} discusses how to exploit multimodal information in domain adaptive pedestrian detection which is not the topic we want to discuss in this paper.

\paragraph{Unsupervised Domain Adaptive Object Detection} It has made a great progress in the challenging unsupervised domain adaptation problem recently \cite{Mnih58,star2020,yaro2015,2017Adversarial,Wang2018Deep,2017Domain}. In the object detection task, most of related methods delve into how to achieve cross-domain alignment between source domain and target domain with different solutions, such as DA-Faster \cite{Chen02}, SW-Faster \cite{Kuniaki}, Region-level Alignment \cite{Zhu16}, CR-DA-DET \cite{Xu06}, ATF \cite{He05}, style transfer based method \cite{Hsu04} and so on. Different from these work, we model the object detection task into a box ranking task to solve the problem of unsupervised false positive suppression without using any domain alignment technique.

\paragraph{Positive-Unlabeled Learning} As mentioned above, domain adaptive object detection is more likely an open-set problem, since the negative instances could be anything around the world, while the positive ones are unique with only domain shift. Therefore, it is naturally connected to positive-unlabeled learning \cite{PUL}. Our proposed \emph{box re-ranking} framework can be regarded as an improved version of positive-unlabeled learning to solve unsupervised false positive suppression problem. To the best of our knowledge, this is the first time we connect positive-unlabeled learning to domain adaptive object detection problem.

%------------------------------------------------------------------------
\section{Results}

\subsection{Experimental Settings}

\paragraph{Datasets and network architecture.} To demonstrate the effectiveness of our proposed method, we build four pedestrian detection domain adaptation tasks. MS COCO \cite{2014Microsoft} is a very large-scale object detection dataset consisting of pedestrian detection, which is taken as source domain dataset in this paper. Following YOLOv3 \cite{yolov3}, we train a YOLOv3 608$\times$608 with backbone Darknet-53 on MS COCO as a strong pre-trained model for unsupervised optimization on downstream target domains. Four challenging pedestrian detection datasets are selected as downstream target domains, including Caltech \cite{Piotr2009Pedestrian}, Cityperson \cite{2017CityPersons}, KITTI \cite{2013Vision} and KAIST \cite{2013Multispectral}. Briefly speaking, Caltech contains 10 hours of video from an urban driving environment, Cityperson records street views across 18 different cities in Germany with various weather conditions, KITTI is a popular urban object detection dataset, and KAIST is a multispectral pedestrian detection dataset with well-aligned visible / themal pairs captured during day and night. Detailed information about training size, validation size, and image resolution on source and target domains are shown in Table.\ref{datasets}. To further verify our proposed method, we also build a more challenging target domain dataset via self-collected data, and we plan to release this dataset later. 

\begin{table}[]
    \newcommand{\tabincell}[2]{\begin{tabular}{@{}#1@{}}#2\end{tabular}}
    \centering
    \begin{tabular}{llccc}
        \specialrule{0.075em}{0pt}{1pt}
        D & Dataset & Train & Val & Resolution\\ 
        \specialrule{0.05em}{1pt}{1pt}
        S & MS COCO \cite{2014Microsoft,yolov3} & 118287 & 5000 & 608$\times$608\\
        \specialrule{0.05em}{1pt}{1pt}
        \multirow{4}{*}{T} & Caltech \cite{Piotr2009Pedestrian} & 4250 & 4024 & 640$\times$480 \\
        & Cityperson \cite{2017CityPersons} & 2975 & 500 & 2048$\times$1024 \\
        & KITTI \cite{2013Vision,2016A} & 3712 & 3769 & 1248$\times$384\\
        & KAIST \cite{2013Multispectral} & 7601 & 2252 &  640$\times$512\\
        \specialrule{0.075em}{1pt}{0pt}
    \end{tabular}
    \caption{Cross-domain pedestrian detection datasets (Domain (D), Source (S), Target (T), Train Size (Train), Validation Size (Val).}
    \label{datasets}
\end{table}

\paragraph{Evaluation metrics. } To describe the performance of false positive suppression, log-average Miss Rate over False Positive Per Image (FPPI) ranging in [$10^{-2}$, $10^0$], which is short for MR in this paper, is utilized as a main evaluation metric. It is also popularly used in many other pedestrian detection work. A good order among true positive boxes and false positive boxes leads to a small MR. Without specific statement, we only determine MR under IoU=0.5.  To further evaluate our proposed method comprehensively, Average Precision (AP) is also presented along with MR as an auxiliary evaluation metric.

\subsection{Implementation details}

We initialize $h=0.95$ and $o=0.4$ to drive seed positive boxes selection and \emph{box number alignment} in a very conservative way. We carry out deep transfer clustering to the boxes with confidence greater than $0.1$. Detailedly, the boxes in the clusters with average confidence greater than 0.95 are set as seed positive boxes. Note that we only carry out deep transfer clustering once before \emph{box re-ranking}. In each repetition for confidence re-prediction, we zero out the confidence of initial boxes generated by the pre-trained model, and match them to the re-predicted boxes generated by the optimized model with maximum IoU (at least greater than 0.3) for confidence updating. Without specific statement, in each repetition, the boxes in top-3 clusters with highest confidence in the negative part are mined back to positive part to drive the next repetition until reaching the condition of \emph{box number alighnment}. The experiments on following cross-domain pedestrian detection datasets share the same implementation configurations. The entire pipeline can be referred to Algorithm.\ref{alg}.

\subsection{Ablation Studies}

\begin{figure}[t]
    \centering
    \includegraphics[width=0.8\columnwidth]{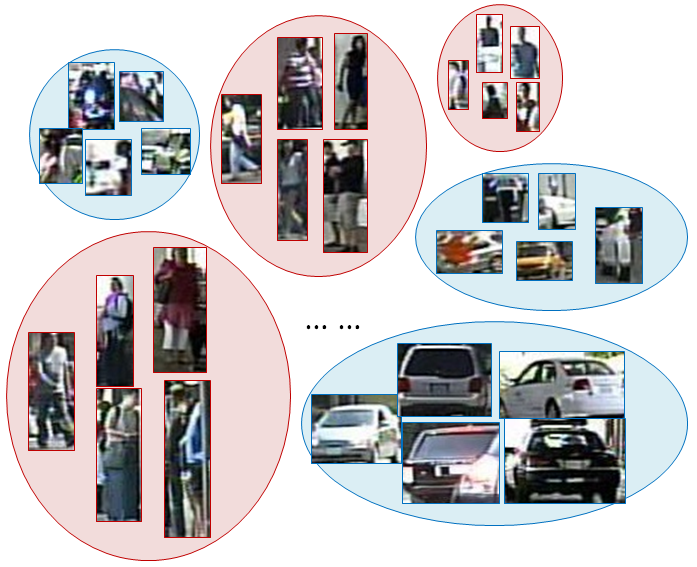} 
    \caption{Visualization of deep transfer clustering on Caltech.} 
    \label{DTC}
\end{figure}

\paragraph{Deep transfer clustering on cross-domain object detection task. } Despite agnostic domain shift, the boxes with similar semantic feature are able to get into the same cluster. What's more, as what we claim above, the scale and aspect ratio information encoded into YOLOv3 feature for box regression are also beneficial for object clustering. As shown in Fig.\ref{DTC}, the boxes with similar semantic, scale and aspect ratio information are well-grouped, which divides true positive and false positive boxes into different clusters.

\begin{figure}[t]
    \centering
    \includegraphics[width=0.8\columnwidth]{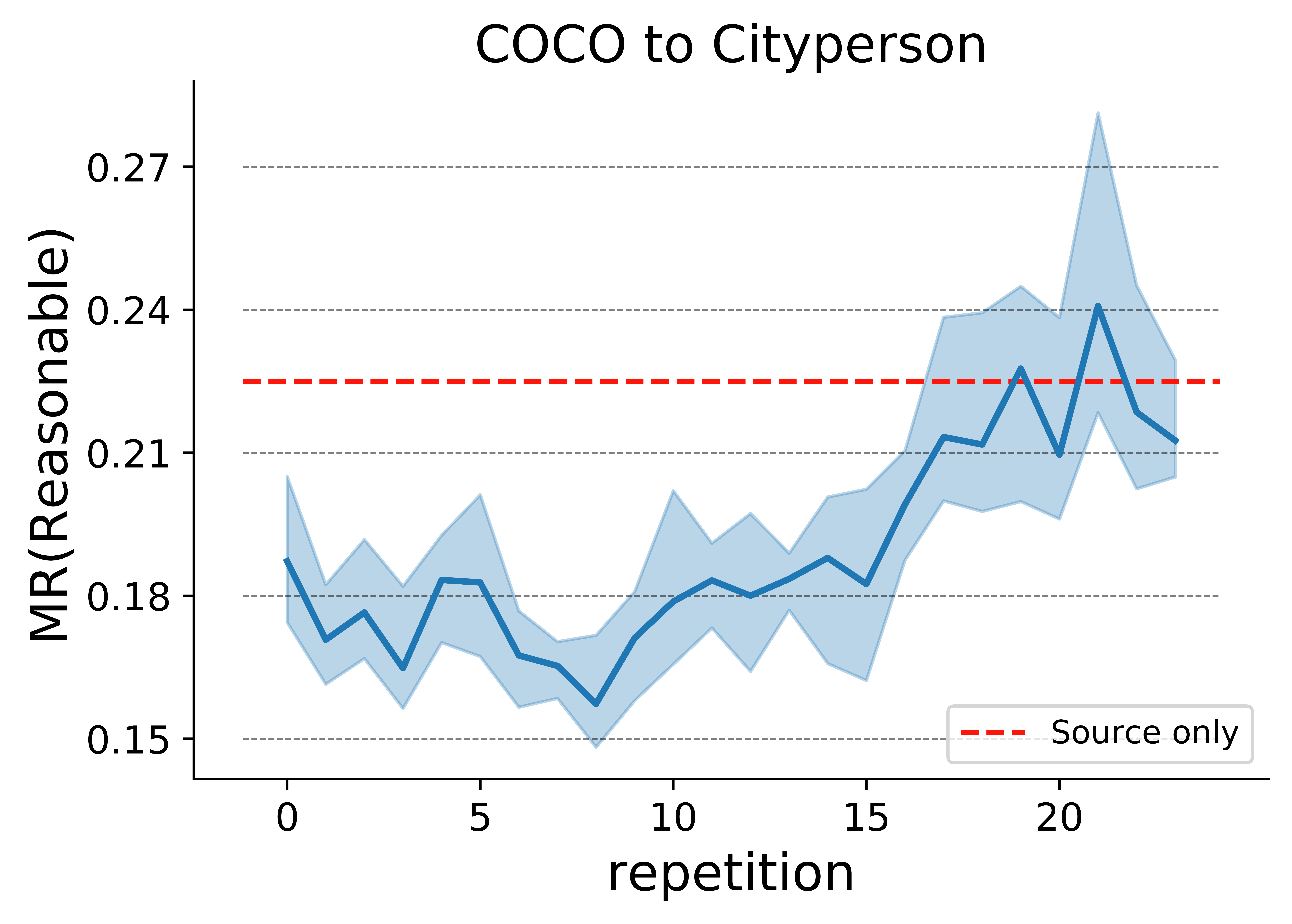} 
    \caption{The varying curves of MR during \emph{box re-ranking}. }
    \label{MR}
\end{figure}

\paragraph{Performance in each repetition during box re-ranking.} To take a deep look at how \emph{box re-ranking} works, we present the varying curve of MR as \emph{box re-ranking} goes on. As shown in Fig.\ref{MR}, MR decreases at first since the order of false positives with higher confidence are swapped with false negatives with lower confidence. As \emph{box re-ranking} goes on, fewer and fewer false negatives are left to be swapped with false positives, resulting in that false positives might be added to positive part for fine-tuning and further leading to MR increasing. This is usually the extremely worst case resulted from over box re-ranking and violates the third prerequisite proposed in section.\ref{prerequisites}. Our proposed \emph{box number alignment} (short for BNA) is a useful self-supervised evaluation metric to avoid this situation. We terminate \emph{box re-ranking} when the box number predicted by the optimized model is comparable as before under a given confidence threshold, namely $o$ in this paper. $o$ is usually used in practical object detection scenarios to output boxes. This operation can prevent our optimized model from over box re-ranking as well as from detection degeneration of true positive. To test BNA to the most extent, we venture to set $o=0.4$ to a very low score in this paper. 

\begin{table}[!t]
	\setlength\tabcolsep{8.5pt}
    \newcommand{\tabincell}[2]{\begin{tabular}{@{}#1@{}}#2\end{tabular}}
    \centering
    \begin{tabular}{lllll}
        \specialrule{0.075em}{0pt}{1pt}
\multirow{2}{*}{Method} & \multicolumn{2}{c}{Reasonable} &  \multicolumn{2}{c}{All} \\
& MR & AP & MR & AP \\
        \specialrule{0.05em}{1pt}{1pt}
Source only & 20.06 & 79.83 & 65.20 & 41.10\\
Oracle &11.25 & 88.18 & 52.47 & 55.22\\
        \specialrule{0.05em}{1pt}{1pt}
Ours (BNA) & 14.76 & 85.37 & 60.54 & 46.41\\
Ours (Best) & 13.57 & 86.34 & 59.89 & 46.72\\
        \specialrule{0.075em}{1pt}{0pt}
    \end{tabular}
    \caption{Results of adaptation from COCO to Caltech.}
    \label{Caltech}
\end{table}
\begin{table*}[t]
    \newcommand{\tabincell}[2]{\begin{tabular}{@{}#1@{}}#2\end{tabular}}
    \centering
    \begin{tabular}{lllllllll}
        \specialrule{0.075em}{0pt}{1pt}
\multirow{2}{*}{Method} & \multicolumn{2}{c}{Reasonable} &  \multicolumn{2}{c}{Bare} & \multicolumn{2}{c}{Partial} & \multicolumn{2}{c}{Heavy}\\
& MR & AP & MR & AP & MR & AP & MR & AP \\
        \specialrule{0.05em}{1pt}{1pt}
Source only & 22.50 & 91.95  & 12.19 & 94.68  & 24.80 & 88.26  & 84.77 & 24.60\\
Oracle & 13.01 & 95.66 & 8.28 & 96.52 & 12.75 & 94.77  & 45.27 & 77.34\\
        \specialrule{0.05em}{1pt}{1pt}
Ours (BNA) & 17.04  & 93.81 & 8.60 & 96.09 & 19.59 & 91.10 & 78.84 & 31.97\\
Ours (Best) & 14.83 & 93.95 & 7.80 & 96.25 & 16.87 & 91.88  & 77.77 & 33.13\\
        \specialrule{0.075em}{1pt}{0pt}
    \end{tabular}
    \caption{Results of adaptation from COCO to Cityperson.}
    \label{Cityperson}
\end{table*}
\begin{table}[!t]
	\setlength\tabcolsep{3pt}
    \newcommand{\tabincell}[2]{\begin{tabular}{@{}#1@{}}#2\end{tabular}}
    \centering
    \begin{tabular}{lllllll}
        \specialrule{0.075em}{0pt}{1pt}
\multirow{2}{*}{Method} & \multicolumn{2}{c}{Easy} &  \multicolumn{2}{c}{Moderate} & \multicolumn{2}{c}{Hard}\\
& MR & AP & MR & AP & MR & AP \\
        \specialrule{0.05em}{1pt}{1pt}
Source only & 38.31 & 51.09 & 50.37  & 49.11 & 60.14 & 42.69\\
Oracle &19.77 & 79.10 & 28.55 & 73.67 & 36.03 & 69.41\\
        \specialrule{0.05em}{1pt}{1pt}
Ours (BNA) & 34.26 &57.14 & 46.46 & 55.50 & 56.85 & 49.55\\
Ours (Best) & 30.04 & 64.15 & 41.85 & 61.13 & 52.97 & 53.88\\
        \specialrule{0.075em}{1pt}{0pt}
    \end{tabular}
    \caption{Results of adaptation from COCO to KITTI.}
    \label{KITTI}
\end{table}
\begin{table}[!t]
	\setlength\tabcolsep{8.5pt}
    \newcommand{\tabincell}[2]{\begin{tabular}{@{}#1@{}}#2\end{tabular}}
    \centering
    \begin{tabular}{lllll}
        \specialrule{0.075em}{0pt}{1pt}
\multirow{2}{*}{Method} & \multicolumn{2}{c}{Day} &  \multicolumn{2}{c}{Night} \\
& MR & AP & MR & AP \\
        \specialrule{0.05em}{1pt}{1pt}
Source only & 39.53 & 72.13 & 46.57 & 60.20\\
Oracle & 26.33 &84.45 & 40.77 & 67.45\\
        \specialrule{0.05em}{1pt}{1pt}
Ours (BNA) & 32.27 & 77.77 & 43.38 & 63.38\\
Ours (Best) &  32.94 & 76.93 & 41.78 & 65.24 \\
        \specialrule{0.075em}{1pt}{0pt}
    \end{tabular}
    \caption{Results of adaptation from COCO to KAIST RGB.}
    \label{KAISTRGB}
\end{table}
\begin{table}[!t]
	\setlength\tabcolsep{8.5pt}
    \newcommand{\tabincell}[2]{\begin{tabular}{@{}#1@{}}#2\end{tabular}}
    \centering
    \begin{tabular}{lllll}
        \specialrule{0.075em}{0pt}{1pt}
\multirow{2}{*}{Method} & \multicolumn{2}{c}{Day} &  \multicolumn{2}{c}{Night} \\
& MR & AP & MR & AP \\
        \specialrule{0.05em}{1pt}{1pt}
Source only & 74.33 & 34.00 & 58.45 & 50.35\\
Oracle & 28.03 & 82.05 & 11.63 & 92.81\\
        \specialrule{0.05em}{1pt}{1pt}
Ours (BNA) & 57.11 & 53.69 & 37.64 & 70.66 \\
Ours (Best) &  54.29 & 57.88 & 32.43 & 75.52 \\
        \specialrule{0.075em}{1pt}{0pt}
    \end{tabular}
    \caption{Results of adaptation from COCO to KAIST Thermal.}
    \label{KAISTTermal}
\end{table}

\begin{figure*}[t]
    \centering
    \includegraphics[width=2.11\columnwidth]{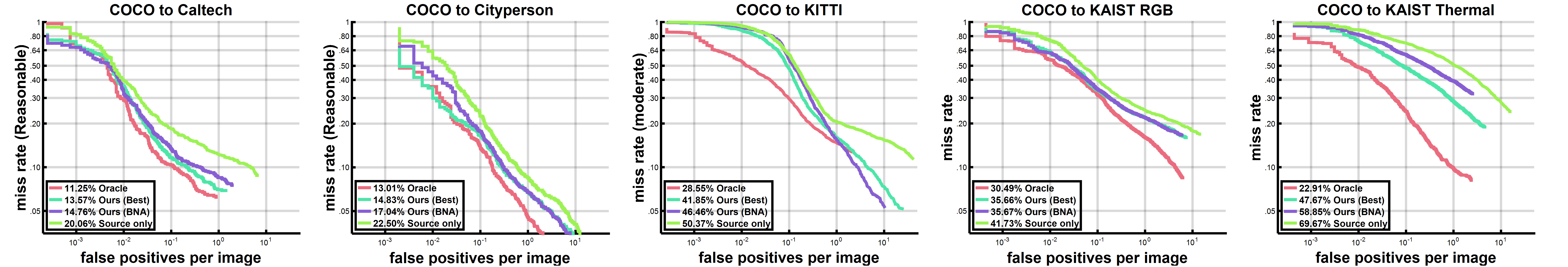} 
    \caption{Performance comparison before and after \emph{box re-ranking}. (miss rate vs. FPPI.)}
    \label{1}
\end{figure*}

\paragraph{Self-supervised vs. supervised evaluation.} We discuss what are the best performances our proposed \emph{box re-ranking} can achieve via supervised evaluation with labeled validation data (best). And we also compare the performances achieved via BNA with supervised evaluation counterparts. As shown in Table.\ref{Caltech},\ref{Cityperson},\ref{KITTI},\ref{KAISTRGB},\ref{KAISTTermal}, our method can decrease MR and increase AP by a large margin compared with source only, the performance directly tested by the pre-trained model. Also, the performance gap between supervised and self-supervised evaluation are relatively small.

\paragraph{Comparison with supervised training.} We use ground-truth labels of training dataset on target domain for supervised fine-tuning. It can reflect the performance upper-bound (oracle) for the unsupervised optimization methods, so as to more precisely evaluate our proposed method. As shown in Table.\ref{Caltech},\ref{Cityperson},\ref{KITTI},\ref{KAISTRGB},\ref{KAISTTermal}, our method can improve the performance in terms of MR and AP. Also, as shown in Table.\ref{KAISTTermal}, our method can also settle the adaptation problem from RGB image to thermal image without effort. In general, there is still exist a performance gap between our method and oracle one. And we mainly impute it to the mis-detection of false negatives instead of the detection of false positives. And our paper only focus on false positive suppression. Nevertheless, based on our proposed false positive suppression framework, we hope to dig out more false negatives which are hard to detect even if we set the confidence threshold near to zero. And we leave this as future work. 

\begin{table}[]
	\setlength\tabcolsep{6pt}
    \newcommand{\tabincell}[2]{\begin{tabular}{@{}#1@{}}#2\end{tabular}}
    \centering
    \begin{tabular}{lllll}
        \specialrule{0.075em}{0pt}{1pt}
\multirow{2}{*}{Method} & \multicolumn{2}{c}{Reasonable} &  \multicolumn{2}{c}{All} \\
& MR & AP & MR & AP \\
        \specialrule{0.05em}{1pt}{1pt}
Ours-cluster (Best) & 13.57 & 86.34 & 59.89 & 46.72\\
Ours-instance (Best) & 15.12  & 85.19  & 61.60  & 44.60 \\
        \specialrule{0.075em}{1pt}{0pt}
    \end{tabular}
    \caption{Performance comparison between instance-based and cluster-based box re-ranking on Caltech.}
    \label{InsClu}
\end{table}
\begin{table}[]
	\setlength\tabcolsep{9pt}
    \newcommand{\tabincell}[2]{\begin{tabular}{@{}#1@{}}#2\end{tabular}}
    \centering
    \begin{tabular}{lllll}
        \specialrule{0.075em}{0pt}{1pt}
\multirow{2}{*}{Method} & \multicolumn{2}{c}{Reasonable} &  \multicolumn{2}{c}{All} \\
& MR & AP & MR & AP \\
        \specialrule{0.05em}{1pt}{1pt}
Source only & 26.73  & 70.17  & 68.71  & 36.89\\
        \specialrule{0.05em}{1pt}{1pt}
Ours (BNA)& 15.46 & 84.33  & 61.28 & 45.37\\
Ours (Best)& 13.95 & 84.08  & 60.41 & 43.72\\
        \specialrule{0.075em}{1pt}{0pt}
    \end{tabular}
    \caption{Performance on the Caltech-Fish dataset by adding strongly hard false positives.}
    \label{fish}
\end{table}

\paragraph{Instance-based vs. cluster-based box re-ranking.} We also carry out an ablation experiment on Caltech dataset to demonstrate the effectiveness of deep transfer clustering in our framework. As shown in Table.\ref{InsClu}, cluster-based box re-ranking performs much better than instance one. The main reason is that the noisy labels in seed positive boxes are cleaned via clustering and the confidence determined in a cluster is more robust for box re-ranking. 

\begin{figure}[t]
    \centering
    \includegraphics[width=1.0\columnwidth]{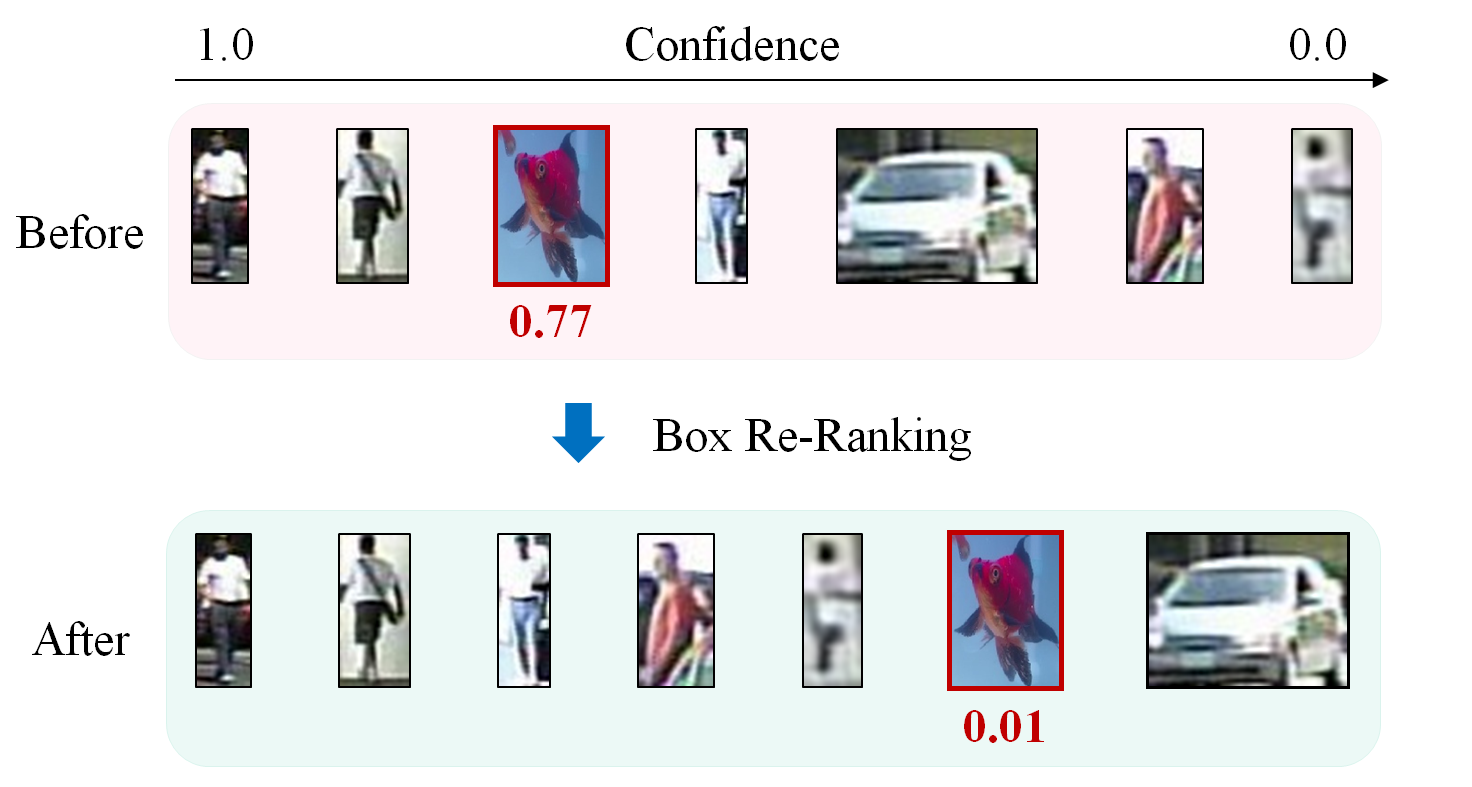} 
    \caption{Illustration of suppressing goldfish via \emph{box re-ranking}. }
    \label{fishresult}
\end{figure}

\paragraph{A more challenging dataset.} The existing public pedestrian datasets are not specifically designed for cross-domain false positive suppression task. To further verify our method, we collect a more challenging dataset in an indoor environment. In this dataset, the false positives are actually goldfish raised indoor as shown in Fig.\ref{fish}, since fish is an unseen object in MS COCO dataset. This is the reason why we call cross-domain false positive suppression is an open-set problem in the above section. As shown in Fig.\ref{fishresult}, the detected goldfish averagely has 0.77 confidence to false-detect into a pedestrian (the highest confidence even goes to 0.99). Since few pedestrians appear in this dataset, we mix it with Caltech dataset (Caltech-Fish dataset) to conduct the following experiment. As shown in Tab.\ref{fish} and Fig.\ref{fishresult}, our method can well-solve this problem. The confidence of detected goldfish is averagely lower to 0.01, while detection capacity of true positive is unchanged. 

\begin{figure}[t]
    \centering
    \includegraphics[width=0.9\columnwidth]{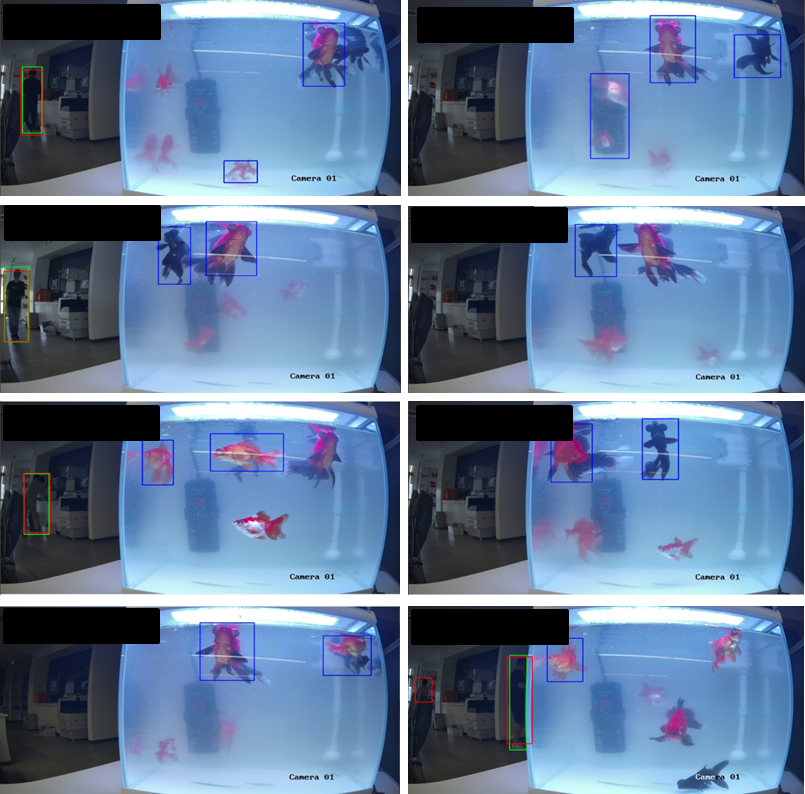} 
    \caption{More strongly hard false positive examples are collected for false positive suppression in cross-domain pedestrian detection. Red, blue and green box separately denote ground-truth, false positive, and true positive box.}
    \label{fish}
\end{figure}

\begin{figure*}[t]
    \centering
    \includegraphics[width=2.1\columnwidth]{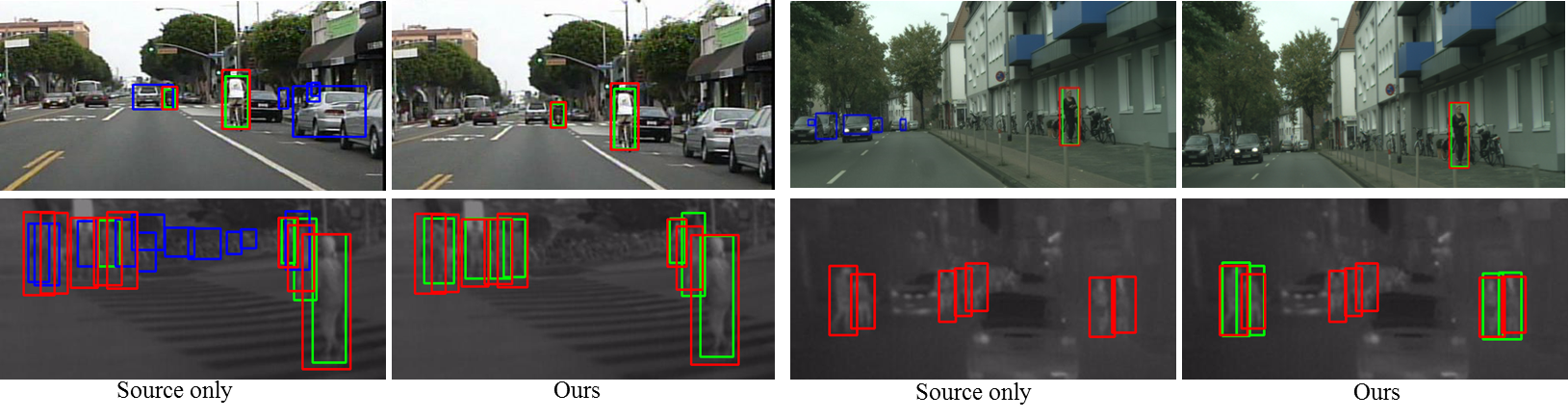} 
    \caption{Visualization of false positive suppression (testing under IoU$=0.5$). Top left: MS COCO to Caltech. Top right: MS COCO to cityperson. Bottom: MS COCO to KAIST Termal. Red, green and blue boxes denote ground-truth, true positive and false positive, respectively.}
    \label{VFPS}
\end{figure*}

\subsection{Discussion}

\paragraph{Visualization of false positive suppression.} Our method claims to re-rank the confidence order of false positive and false negative. In this way, with comparable box number to be detected, we can suppress the false positive while mining back the considerable part of false negative. As shown in Fig.\ref{VFPS}, we can effectively suppress false positive especially those just produced by agnostic domain noise with different semantic information from pedestrians. Also, it is worth mentioning that our method can also rectify the bounding boxes with greater IoU matched with ground truth boxes, which may be owe to the Law of Large Numbers.

\paragraph{Training cost.} We try to keep the training configuration exactly the same so as to compare the performance among different repetitions fairly during \emph{box re-ranking}. We always initialize the model with pre-trained weights from source domain, and fine-tune the model on target domain by only 5 epochs, since no source domain data join training and thus avoiding domain conflict between source and target domain during training. Different datasets require different rounds of repetition to finish \emph{box re-ranking}. Roughly estimating from the experiments conducted above, about 10 repetitions are sufficient to terminate \emph{box re-ranking}, which means about 50 epochs are required for fine-tuning totally. In practical scenarios without the need for fair performance comparison, this training cost can be further optimized.

\begin{table}[]
    \centering
    \begin{tabular}{lc}
        \specialrule{0.075em}{0pt}{1pt}
Method & AP of Car \\
        \specialrule{0.05em}{1pt}{1pt}
Source only & 36.4\\
Oracle & 58.5 \\
        \specialrule{0.05em}{1pt}{1pt}
DA-Faster \cite{Chen02} & 38.5\\
SW-Faster \cite{Kuniaki} & 37.9\\
Noise Labeling \cite{Khodabandeh17} & 43.0\\
DA-Detection \cite{Hsu04} & 43.9\\
AT-Faster \cite{He05} & 42.1\\
        \specialrule{0.05em}{1pt}{1pt}
Ours & 44.2\\
        \specialrule{0.075em}{1pt}{0pt}
    \end{tabular}
    \caption{Results of adaptation from KITTI to Cityscapes.}
    \label{K2C}
\end{table}
\begin{table}[]
    \centering
    \begin{tabular}{lc}
        \specialrule{0.075em}{0pt}{1pt}
Method & AP of Car \\
        \specialrule{0.05em}{1pt}{1pt}
Source only & 33.7\\
Oracle & 58.5 \\
        \specialrule{0.05em}{1pt}{1pt}
DA-Faster \cite{Chen02} & 38.5\\
Noise Labeling \cite{Khodabandeh17} & 43.0\\
AT-Faster \cite{He05} & 42.1\\
        \specialrule{0.05em}{1pt}{1pt}
Ours & 44.0\\
        \specialrule{0.075em}{1pt}{0pt}
    \end{tabular}
    \caption{Results of adaptation from Sim10k to Cityscapes.}
    \label{S2C}
\end{table}

\paragraph{Future work.} \emph{Box number alignment} in this paper is proposed for false positive suppression. Beyond false positive suppression, more effective unsupervised evaluation metrics should be developed in the future for unsupervised learning with less prerequisite dependency.

\subsection{Extension to General Benchmarks}

Since cross-domain unsupervised false positive suppression is a new task we propose in this paper, no previous work is appropriate to discuss with our proposed \emph{box re-ranking} framework quantitatively. To further verify our proposed method, we also compare our proposed method with other state-of-the-arts in two general domain adaptive object detection datasets, including the adaptation from KITTI to Cityscapes \cite{Cordts47} and the adaptation from Sim10k \cite{Johnson-Roberson46} to Cityscapes. Following other work \cite{Chen02,Kuniaki,Khodabandeh17,Hsu04,He05}, these two adaptation datasets use Faster-RCNN \cite{Ren2017Faster} for car detection and take Average Precision (AP) as the main evaluation metric. Different from YOLOv3, we use ROI pooling feature of each box for deep transfer clustering in Faster-RCNN. As shown in Table.\ref{K2C} and \ref{S2C}, our proposed method outperforms other competing methods. 

%-----------------------------------------------------------------------
\section{Conclusion}
In this paper, a new task, named unsupervised false positive suppression for domain adaptive pedestrian detection, is proposed. In order to bypass this challenge elegantly, we model an object detection task into a box ranking task among true positive and false positive, and further transform false positive suppression problem into a \emph{box re-ranking} problem. Under this modeling, false positive is able to be suppressed effectively without any manual annotation, which has been supported by extensive experiments carried out in this paper. Our proposed framework is very practical to upgrade object detection model into scene-specific one. We hope our method can bring rich inspirations to the community of unsupervised learning.

\begin{figure*}[!htp]
    \centering
    \includegraphics[width=2.0\columnwidth]{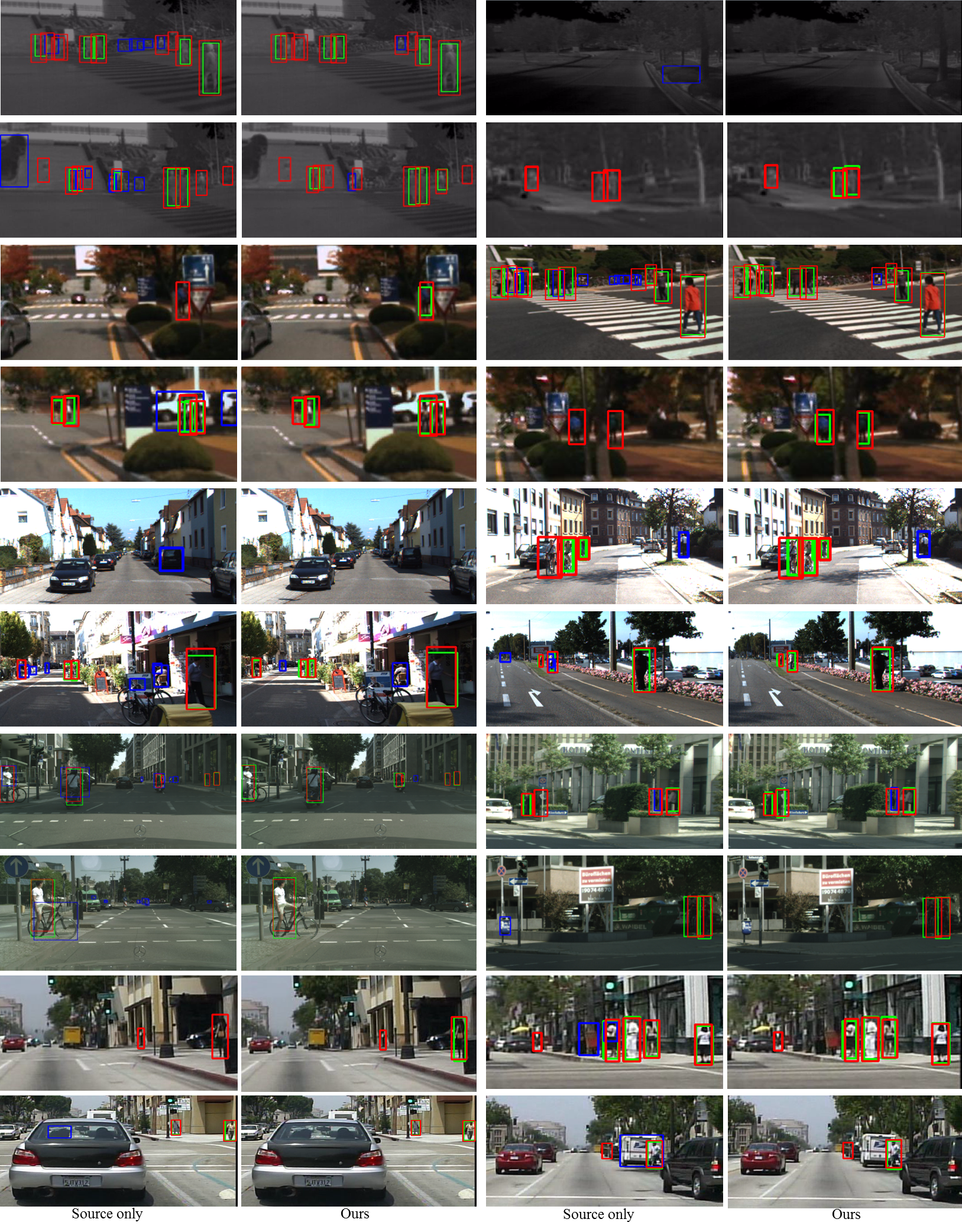} 
    \caption{More visualization of false positive suppression on public-released datasets (IoU=0.5), including KAIST, KITTI, Cityperson and Caltech. Under the constraint of our proposed box number alignment, the more false positives are suppressed, the more false negatives will be mined back. Red, blue and green box separately denote ground-truth, false positive, and true positive box. A relatively small part of blue boxes are resulted from mis-annotation of true positives.}
    \label{morevisualization}
\end{figure*}

{\small
\bibliographystyle{ieee_fullname}
\bibliography{egbib}
}

\end{document}